\documentclass{article}
\usepackage[utf8]{inputenc}
\usepackage{amsmath}
\usepackage{amssymb}
\usepackage{graphicx}
\usepackage{hyperref}
\usepackage{cite}
\usepackage{tabularx}
\usepackage{float}
\usepackage{authblk}

\title{Trustworthy AI: Securing Sensitive Data \\ in Large Language Models}

\author[1]{Georgios Feretzakis}
\author[1]{Vassilios S. Verykios}
\affil[1]{Big Data Analytics and Anonymization lab, School of Science and Technology, Hellenic Open University, Patras, Greece \\ 
\par \texttt{georgios.feretzakis@ac.eap.gr} \\ \texttt{verykios@eap.gr}}

\date{}
\begin{document}

\maketitle

\section*{Abstract}
Large Language Models (LLMs) have transformed natural language processing (NLP) by enabling robust text generation and understanding. However, their deployment in sensitive domains like healthcare, finance, and legal services raises critical concerns about privacy and data security. This paper proposes a comprehensive framework for embedding trust mechanisms into LLMs to dynamically control the disclosure of sensitive information. The framework integrates three core components: User Trust Profiling, Information Sensitivity Detection, and Adaptive Output Control. By leveraging techniques such as Role-Based Access Control (RBAC), Attribute-Based Access Control (ABAC), Named Entity Recognition (NER), contextual analysis, and privacy-preserving methods like differential privacy, the system ensures that sensitive information is disclosed appropriately based on the user’s trust level. By focusing on balancing data utility and privacy, the proposed solution offers a novel approach to securely deploying LLMs in high-risk environments. Future work will focus on testing this framework across various domains to evaluate its effectiveness in managing sensitive data while maintaining system efficiency.

\noindent \textbf{Keywords:} Large Language Models, Trust Mechanisms, Sensitive Information, Role-Based Access Control, Attribute-Based Access Control, Data Privacy, Privacy-Preserving Techniques, Named Entity Recognition, Differential Privacy, AI Ethics.

\section{Introduction}

\subsection{Background}
\textbf{Overview of Large Language Models (LLMs) and Their Significance}

Natural Language Processing (NLP) has seen phenomenal progress, driven by increasingly complex LLMs, which extend the boundary of how much machines can actually understand and generate text. They are now powerful enough to generate coherent, contextually relevant text that feels human-like. Among recent releases are GPT-4o, o1 series by OpenAI, Gemini 1.5 Pro from Google DeepMind, Claude 3.5 Sonnet by Anthropic, and LLaMA 3.1 by Meta AI.
These are the foundation of models for applications including advanced conversational agents, real-time translation, text summarization, and question-answering systems providing state-of-the-art performance in industries such as healthcare, legal services, finance, and education. By leveraging vast amounts of data, LLMs can address increasingly complex problems, further pushing the frontier of service development based on AI and personalizing user experiences across various domains.

This progress builds on earlier advancements in NLP, where LLMs like BERT from Google \cite{devlin2019bert}, GPT-3 from OpenAI \cite{brown2020gpt3}, and RoBERTa from Facebook \cite{liu2019roberta} were instrumental in revolutionizing tasks such as machine translation, summarization, question answering, and conversational agents. These earlier models laid the groundwork for the powerful systems we see today, contributing to the remarkable state-of-the-art performance achieved in various NLP applications.

In other words, LLMs are deep neural networks with billions of parameters trained on vast amounts of data that include all types of text available on the World Wide Web. They are capable of generating coherent, contextually relevant, fluent output. This has made them a critical component in a wide array of applications for healthcare, finance, customer service, and education \cite{kalyan2020sec} \cite{sun2019bertclassification}. Examples include deploying them to automate customer support through chatbots, interpreting clinical notes to provide diagnostic help in medicine, and improving educational tools to offer more personalized learning experiences.

LLMs also capture these complex linguistic patterns of the language and contextual relations crucial in it; hence, they are able to execute many activities that were earlier considered very challenging for AI systems \cite{vaswani2017attention}. They have contributed to automating routine tasks, enhancing user experiences through personalization, and driving innovation into AI-powered services.

\textbf{The Growing Concern over Sensitive Information Management in AI}

However, simultaneously, LLMs have become a growing concern with regard to the handling of sensitive information \cite{bommasani2021foundation}. Due to the large-scale nature of their training datasets, LLMs may inadvertently contain personal, confidential, or proprietary information. Thus, such LLMs can unintentionally produce answers that disclose such data \cite{mcmillan2020privacylm}. Research has shown that LLMs can memorize and regurgitate pieces of their training data, even if it is sensitive personal information such as names, addresses, or unique identifiers \cite{carlini2019secretsharer} \cite{lehman2021bertclinical}. This presents serious risks to privacy and security when LLMs are deployed in applications accessed by people or in situations where data confidentiality is of utmost importance.

One of the critical issues emerging today is that if such LLMs are employed within systems dealing with sensitive information—legal, medical, or financial services—the occurrence of unauthorized leakage of confidential data is probably inevitable \cite{shokri2015deepprivacy}. Regulations such as the General Data Protection Regulation (GDPR) \cite{eu2016gdpr} and the Health Insurance Portability and Accountability Act (HIPAA) \cite{hipaa1996} prescribe the most stringent norms of handling and protecting personal or sensitive data, hence demanding the already urgent need for the execution of effective methods to prevent data breaches.

There is now increasing attention to developing methods that avoid the memorization and leakage by the LLM of information it should not, for example, through training with differential privacy \cite{abadi2016deepprivacy} \cite{carlini2021extracting}. This security mechanism can be researched in more detail within this thesis, to further develop well-performing AI systems with certain privacy enhancements, balancing the benefits of LLMs while protecting sensitive information.

\subsection{Problem Statement}
\textbf{Challenges in Preventing Unauthorized Disclosure of Sensitive Information by LLMs}

While LLMs have unparalleled capabilities to generate text-like human language, this often involves considerable risks in terms of possible leakage or disclosure of private information. The issue is that LLMs can memorize and produce parts of their training datasets, which include PII data and confidential information at relevant questioning \cite{carlini2019secretsharer} \cite{carlini2021extracting}. This presents a significant challenge, considering that many LLMs are trained on extremely large sets of internet-scraped data, which might contain sensitive or private information \cite{carlini2021extracting} \cite{zhang2020deepmutuallearning}.

Attackers can exploit LLMs through model inversion or other methods to extract sensitive data \cite{fredrikson2015modelinversion} \cite{song2017memoryoverload}. For example, an attacker might incorporate a series of crafted prompts to extract social security numbers, credit card information, or private communications that it may have inadvertently learned from the model's training dataset \cite{carlini2021extracting} \cite{ateniese2015smartmachines}. This problem increases, given the black-box nature of most LLMs: their internal mechanisms have remained rather opaque. It gives very little insight into whether one can predict or control what the model may reveal \cite{bender2021stochasticparrots}.

Also, since LLMs are probabilistic and generating text is a probability in itself, auditing and controlling its outputs are very hard to achieve \cite{papernot2016distillation}. Classic content filtering methods such as keyword blacklists do not work since the same sensitive information can be expressed using paraphrased or contextually changed language \cite{krishna2021paraphrase}. Besides, it may generate sensitive information as a part of its normal operation without needing any malicious prompting \cite{li2018deleteretrievegenerate}.

The sensitivity lies in the balance between utility to LLMs and concerns about sensitive information protection \cite{shokri2015deepprivacy}. Over-constraining the output will lead to a model's lowered effectiveness and limited applications for legitimate use. In contrast, this will enhance the risks of data breaches and non-compliance with privacy regulations if adequate checks are not in place \cite{veale2017fairml}. This will require a sophisticated balance, wherein the sensitivity level of the information being processed and generated by the large language models will be dynamically assessed and managed.

\textbf{Limitations of Current Approaches in Managing Sensitive Data}

General methods of preventing unauthorized disclosure of sensitive information by LLMs include data sanitization, differential privacy, and output filtering \cite{abadi2016deepprivacy}, \cite{veale2017fairml}, \cite{lyu2020federatedthreats}. Nevertheless, these methods have some important limitations that impair their potential.

Data sanitization is the process of removing sensitive information from the training dataset before models are trained \cite{thudumu2020datasanitization}. This approach in theory can decrease the probability of the model assimilating and reproducing sensitive data, but in practice is often infeasible given the vast amount of training data and the difficulty in recognizing all forms of sensitive content \cite{sweeney2002kanonymity}, \cite{elemam2011reidentification}. Automated sanitization may miss context-specific or subtle forms of sensitive information, while manual sanitization of billions of words in datasets becomes practically impossible \cite{elemam2011reidentification}.

Differential Privacy adds mathematical noise to the training process, ensuring that the model cannot learn details about individual data points and is restricted from memorizing such details \cite{abadi2016deepprivacy}, \cite{dwork2014differentialprivacy}. But then again, though differential privacy offers good theoretical guarantees against data extraction attacks, it degrades the performance of the model in complex language tasks \cite{dwork2014differentialprivacy} \cite{tramer2022accuracy}. Besides, differential privacy on a scale needed for LLMs has high computational requirements and may not be practical in most real-world scenarios \cite{tramer2022accuracy}.

Output filtering inspects the model's outputs, being applied to discard or modify sensitive content before it is delivered to the user \cite{gehrmann2019gltr}. This scheme has difficulty in precisely identifying all occurrences of sensitive information without a high false positive rate \cite{henderson2018ethicaldialogue}. These nuances in contextualization and the model's ability to output semantic variations of the same expression make it even harder for filters to identify every instance of sensitive data disclosure \cite{wallace2019universaladversarial}. Moreover, experienced users will generally find ways to bypass output filters through skillful prompting techniques \cite{wallace2019universaladversarial}.

Most of these approaches tend to also be myopic in terms of the trust level of the user interacting with the LLM \cite{bauer2009accesscontrolchallenges}. They invariably follow a one-size-fits-all policy that does not make a distinction for users based on roles, permissions, or trustworthiness \cite{alam2008trustcomputing}. This inability to discriminate between users can result in either legitimate access for trusted users not being possible or unauthorized disclosure to untrusted users.

In other words, the realization of sensitive information treatment cannot be achieved by any state-of-the-art methods at the current stage of LLMs. This calls loudly for more active and fine-grained methods that embed mechanisms of trust into the very operation of LLMs, allowing responses to be shaped by the user's trust level to fortify security without unduly eroding the utility of the model.

\subsection{Objectives}
\textbf{To Propose a Framework for Embedding Trust Mechanisms in LLMs}

This chapter generally proposes a framework that embeds mechanisms of trust inside LLMs. By incorporating trust management inside LLMs, this framework controls sensitive information sharing based on the end user's level of trust when interacting with such a system. This directly addresses the gaps in previous approaches, which almost never account for the relevant level of trust from users \cite{bauer2009accesscontrolchallenges}, \cite{alam2008trustcomputing}.

In essence, the proposed framework extends RBAC and ABAC ideas within the context of LLMs \cite{bauer2009accesscontrolchallenges}, \cite{hu2014abac}. It defines levels of user trust and associates the correct level of permission concerning sensitive information disclosure. The LLM is then allowed dynamic adjustments in its responses based on the user's permission level.

This paradigm supports utility and effectiveness in LLMs while upholding data safeguards. This would make AI systems more secure in sensitive areas where protection of data is demanded, such as in health, finance, and legal services \cite{shokri2015deepprivacy}. The framework will also consider alignment with legal provisions like GDPR and HIPAA, ensuring conformance of the deployment of LLMs with data protection requirements \cite{eu2016gdpr}, \cite{hipaa1996}.

\textbf{To Develop Methods for Automatic Detection and Control of Sensitive Information Disclosure Based on User Trust Levels}

The second involves the design of methods that will automatically detect and control sensitive information disclosure in LLM outputs, informed by the user's trust level in the model. Such development involves formulating an algorithm and techniques capable of letting the LLM perceive, with its response, what could potentially be sensitive content, and make runtime decisions on disclosure, modification, or retention \cite{lample2016ner}, \cite{joulin2017efficienttext}. According to this purpose, state-of-the-art NLP approaches such as Named Entity Recognition (NER) \cite{lample2016ner}, text classification \cite{joulin2017efficienttext}, and content filtering \cite{gehrmann2019gltr} will be utilized. These can identify sensitive data, such as personal identifiers, confidential corporate information, and proprietary content, within model outputs. The methods will incorporate user trust level in decision-making, ensuring that only users with proper authorization access certain details \cite{bauer2009accesscontrolchallenges}, \cite{hu2014abac}. For instance, a high-trust user may receive responses that are more elaborate, containing sensitive data, while a low-trust user will receive generalized or redacted information.

The work will also explore the use of differential privacy \cite{abadi2016deepprivacy} and other disclosure limitation tools to reduce the risks of sensitive data disclosure by the model itself. These techniques will fit into the overall framework for enhancing privacy and security in LLMs while minimizing performance costs \cite{abadi2016deepprivacy}, \cite{tramer2022accuracy}.

Overall, this research will design a strong and adaptable system that prevents unauthorized access to sensitive data while ensuring LLMs work effectively across different applications.

\section{Literature Review}

\subsection{Large Language Models and Their Limitations}

\textbf{Overview of LLM Architectures and Functionalities}

To date, Large Language Models (LLMs) have significantly advanced the field of Natural Language Processing (NLP) due to their exemplary capabilities in understanding and producing text that mimics human communication \cite{devlin2019bert}. These models use deep learning frameworks to find complex patterns and contextual relationships in language developed from large datasets. The Transformer model by Vaswani et al. \cite{vaswani2017attention} provides a basic architecture for LLMs using self-attention to model text dependencies of long-range much better than previous recurrent architectures.

Prominent examples of LLMs are BERT, GPT, RoBERTa, T5, XLNet, and Meta's series of LLaMA. BERT is an abbreviation for Bidirectional Encoder Representations from Transformers. Devlin et al. \cite{devlin2019bert} proposed this, conditioning on both left and right context to capture deep bidirectional representations. It is pre-trained on masked language modeling and next-sentence prediction tasks, therefore performing really well in a wide range of NLP tasks after fine-tuning.

The best-known GPT series by OpenAI are GPT-2 \cite{radford2019unsupervisedmultitask} and GPT-3 \cite{brown2020gpt3}. They are autoregressive models that predict the next word in a sequence. GPT-3 is the largest, with 175 billion parameters, and has achieved impressive scores on few-shot learning and text generation. RoBERTa is an improved version of BERT proposed by Liu et al. \cite{liu2019roberta}. RoBERTa works better than the above-described models by training on more data and omitting the objective of next sentence prediction.

Text-to-Text Transfer Learning, as proposed by Raffel et al. in T5 \cite{raffel2020t5}, generalizes all NLP tasks into a unified task for easier implementation of transfer learning. XLNet, by Yang et al. \cite{yang2019xlnet}, proposes a permutation-based training objective to model effective bidirectional contexts, which avoids the pretraining constraints of masked language models.

Foundation—LLaMA (Large Language Model Meta AI) \cite{touvron2023llama} has 7 and 65 billion parameters going to Meta AI. LLaMA is engineered for being high-performance with high efficiency; one order of magnitude smaller in size than other bulk models, it still achieves the highest performance. In July 2023, Meta AI released a follow-up to LLaMA, called LLaMA 2 \cite{touvron2023llama2}, which contains 7, 13, and 70 billion parameter models. These are models trained on about 2 trillion tokens—much more than their predecessors—and outperform them on some tasks like reasoning, coding, and instruction following \cite{touvron2023llama2, metaai2023llama2}.

The important features of LLaMA 2 include an expanded training dataset, architecture improvements including increased context length, now up to 4,096 tokens, and improved multilingual supporting more than 20 languages \cite{touvron2023llama2} \cite{ramaswamy2023scalinglaws}  \cite{goyal2021flores}. Furthermore, in the case of LLaMA 2, Meta has focused on responsible AI development by embedding safety into the model and providing responsible use of resources \cite{bommasani2021foundation, touvron2023llama2}.

At this moment, the state-of-the-art base models include LLaMA 3 \cite{metaai2024llama3}, GPT-4 \cite{openai2023gpt4}, and GPT-4o \cite{openai2024gpt4o}, each taking the performance of language models one step ahead. This is further developed with this line: The LLaMA 3 from Meta; the system has 8B and 70B parameter models, with a planned 405B parameter model. To date, LLaMA 3 was trained on extended data to over 15 trillion tokens, seven times more than what was used for training LLaMA 2. This also expanded multilingual features for more than 30 languages and now covers four times more code-related data. Architectural enhancements further include a new tokenizer with 128 and 256 token capabilities, grouped query attention stepping up processing, and sequence handling to maximize management to 8,192 in LLaMA 3. Meta continues its commitment to the responsible development of AI as it rolls out LLaMA Guard 2 and Code Shield, tools designed to shore up trust and safety in model deployment. Early benchmark results show LLaMA 3 greatly outcompetes other open-source models on understanding language, reasoning, and coding.

Another huge leap in LLM technology was made after the release of GPT-4 in March 2023. Generally speaking, GPT-4 is better in functionality than its predecessor, GPT-3.5, since GPT-4 can handle up to 32,768 tokens and provides much better performance on tests of both academic and professional character. More interestingly, increases in reasoning and creativity introduce steerability that allows requests for style and behavior from the users. Of course, GPT-4 does keep biases, and at times it might offer nothing but wrong or ``hallucinated'' information.

The newest addition, GPT-4o, released in May 2024, went one step beyond GPT-4 by adding multimodal input; the newest one processes text, audio, and images. This makes GPT-4o much more applicable across a larger range of applications, given that real-time voice interactions and visual content analysis have been added. Moreover, GPT-4o will run faster and cheaper than GPT-4, meaning it can reach many more users, for free and paying. New features such as support for multiple languages, immediate translation, the development of a desktop application, and easy integration into user workflows make it an indispensable model in many professions. Safety is designed into GPT-4o to ensure better enforcement of trust and safety features compared to previous models and, as such, alleviate some of the ethical issues.

These models are pre-trained first on huge text corpora and further fine-tuned on specific downstream tasks, including text classification \cite{sun2019bertclassification}  \cite{joulin2017efficienttext}, question answering \cite{metaai2023llama2}, machine translation \cite{goyal2021flores}, text summarization \cite{ramaswamy2023scalinglaws}, and conversational AI \cite{bommasani2021foundation}. Only recently have new computational resources and optimization techniques enabled the empirical exploration of scaling up LLMs in a way that consistently improves their performance across tasks \cite{kaplan2020scalinglaws}. At the same time, with these advantages of scaling, new challenges and limitations also arise.

\textbf{Discussion of Issues Related to Information Leakage}

While doing so, LLMs run major risks of information leakage in sensitive or private data in the generated outputs from the training corpus \cite{carlini2019secretsharer, abadi2016deepprivacy}. This presents serious privacy risks, as many models are trained on datasets with no explicit consent for personal or proprietary information.

A significant issue involved is the memorization of training data. Large Language Models can indeed memorize infrequent or unique sequences contained in the training dataset, which later could be verbatim reproduced when a specific prompt is given \cite{carlini2019secretsharer, carlini2021extracting}. This poses problems when the memorized information involves PII or confidential data. Attackers may carry out extraction attacks by crafting inputs that get the model to reveal sensitive information \cite{carlini2021extracting, fredrikson2015modelinversion}. Such attacks would exploit the model's ability to memorize some inputs and then reproduce them.

Traditional mitigation techniques such as anonymization or filtering can only do so much since models may infer or reconstruct sensitive information due to pattern leakage in the data itself \cite{elemam2011reidentification, thudumu2020datasanitization}. Moreover, it is considered computationally infeasible to apply successful methods of privacy preservation on the scale required by LLMs \cite{tramer2022accuracy}. Methods like differential privacy aim to limit the influence of a single datapoint on model parameters \cite{abadi2016deepprivacy, dwork2014differentialprivacy}; however, finding a balance between privacy and utility remains an open challenge \cite{tramer2022accuracy}.

Recent works have pointed out these subjects. Carlini et al. \cite{carlini2021extracting} showed that GPT-2 and GPT-3 models, among others, could involuntarily leak sensitive personal information kept inside their training data. Lehman et al. \cite{lehman2021bertclinical} observed that BERT models pre-trained on clinical notes can leak sensitive patient information; this raises concern about their use in possible medical environments.

Most promising solutions to achieve the above include mechanisms for strict data curation and consents that forbid training on unauthorized sensitive information \cite{sweeney2002kanonymity, elemam2011reidentification}, ways to embed user trust levels and access controls into the LLM systems, and ways to enable dynamic management of the disclosure of sensitive information \cite{bauer2009accesscontrolchallenges, hu2014abac, zhang2015privacypreserving}. However, numerous issues remain: performance tradeoffs and bypassing by the user with rephrased or alternative prompts are still possible \cite{krishna2021paraphrase, wallace2019universaladversarial}.

Information leakage in LLMs is of great concern because increasingly such models will be finding applications in sensitive data environments. There is hence the urgent need for sophisticated solutions capable of impeding unauthorized disclosure, all without affecting model performance. Building mechanisms of trust and creating robust techniques for preserving privacy are what become necessary for responsible deployment.

\subsection{Trust Mechanisms in Computing}

\textbf{Existing Trust Models (e.g., RBAC, ABAC)}

Trust mechanisms are foundational in computing systems to ensure secure and appropriate access to resources. Two predominant models used to manage and enforce trust are Role-Based Access Control (RBAC) and Attribute-Based Access Control (ABAC).

Role-Based Access Control (RBAC) is an access control paradigm where permissions are associated with roles, and users are assigned to these roles, thereby acquiring the permissions of the roles \cite{sandhu1996rbacmodels}. This model simplifies management by grouping permissions into roles that reflect organizational job functions. RBAC is widely adopted in enterprises due to its straightforward approach to access management.

Key components of RBAC include:
\begin{itemize}
    \item Users: Individuals who need access to system resources.
    \item Roles: Defined job functions within an organization.
    \item Permissions: Approval to perform certain operations.
    \item Sessions: Instances where users activate a subset of roles.
\end{itemize}

RBAC operates on the principle of least privilege, ensuring that users have only the access necessary to perform their duties \cite{ferraiolo2003rbac}. It is effective in environments where user responsibilities and permissions are relatively static.

Attribute-Based Access Control (ABAC) extends the capabilities of RBAC by using attributes as the basis for access decisions \cite{hu2014abac, yuan2005abac}. In ABAC, rights to access are provided via policies, written using a combination of the attributes of users, resources, actions, and their environment. Attributes might include such examples as user characteristics (e.g., department, clearance level), resource properties (e.g., the sensitivity of data), action type (e.g., read vs. write), and environmental conditions (such as time or location). ABAC provides better flexibility and granularity, suitable for dynamic environments where access requirements may change frequently \cite{jin2012abacmodel}. It allows the creation of complex policy definitions that can adapt to different situations without requiring new role definitions or modification of existing roles.

\textbf{Comparison and Integration}
While RBAC offers simplicity and ease of administration, ABAC provides the flexibility needed for complex and dynamic access control scenarios \cite{kuhn2010rbacattributes}. There is a growing interest in integrating RBAC and ABAC to leverage the benefits of both models. Such integration aims to enhance scalability and manageability while providing granular access control \cite{jin2012abacmodel, kuhn2010rbacattributes}.

\textbf{Trust in Human-Computer Interaction}
Trust in human-computer interaction (HCI) is critical for user acceptance and effective use of technology \cite{corritore2003onlinetrust, mcknight2001trustdistrust}. It influences how users perceive, engage with, and rely on computing systems, particularly those that handle sensitive information or perform critical functions.

\textbf{Factors Influencing Trust in HCI:}
\begin{enumerate}
    \item \textbf{Reliability:} Consistent and dependable system performance builds user trust \cite{mayer1995trustmodel}. Systems that frequently crash or produce errors can erode confidence.
    \item \textbf{Security and Privacy:} Assurance that personal data is protected is essential for trust \cite{mcknight2001trustdistrust, belanger2011privacy}. Users are more likely to trust systems that demonstrate robust security measures and transparent privacy policies.
    \item \textbf{Usability:} Intuitive interfaces and user-friendly designs enhance trust by making systems accessible and easy to navigate \cite{nielsen1994usabilityengineering}.
    \item \textbf{Transparency and Explainability:} Systems that provide clear explanations for their operations and decisions help users understand and trust the technology \cite{tintarev2007explanationsurvey}.
    \item \textbf{Responsiveness:} Timely and appropriate responses to user inputs contribute to a positive user experience and increased trust \cite{shneiderman2005ui}.
\end{enumerate}

\textbf{Trust Challenges in AI Systems}
With the rise of artificial intelligence and machine learning, new challenges have emerged in establishing trust \cite{lee2004trustinautomation, hoff2015trustinautomation}. AI systems often operate as "black boxes," making it difficult for users to understand how decisions are made. Concerns about algorithmic bias, lack of transparency, and potential errors can undermine trust in AI technologies.

\textbf{Building Trust in AI Systems:}
\begin{itemize}
    \item \textbf{Explainable AI (XAI):} Developing AI models that can provide explanations for their decisions enhances transparency and trust \cite{gunning2019xai}. XAI helps users understand the reasoning behind AI outputs.
    \item \textbf{Ethical Design:} Incorporating ethical principles, such as fairness and accountability, ensure that AI systems operate responsibly \cite{dignum2018ethicsai, venkatesh2003useracceptance}. Addressing ethical considerations proactively can mitigate trust issues.
    \item \textbf{User-Centered Design:} Involving users in the design process helps align the system with user needs and expectations \cite{nielsen1994usabilityengineering, venkatesh2003useracceptance}. Feedback mechanisms and iterative design improve usability and trustworthiness.
    \item \textbf{Security Measures:} Implementing robust security protocols protects against unauthorized access and data breaches, reinforcing user trust \cite{belanger2011privacy}.
\end{itemize}

\textbf{Implications for LLMs}
Clearly, trust is an element of great importance in interacting and accepting Large Language Models (LLMs) by users in \cite{liao2020explainabledesign}. Applications regarding sensitive information, such as virtual assistants, customer service bots, and systems that support decision-making, are being led by large language models. The user needs to trust that the system will result in trusted, relevant, and appropriate information.

Building trust mechanisms for LLMs through integration of mechanisms from access control models such as RBAC and ABAC, to manage information disclosure according to the user's trust level \cite{zhang2019abacapis}. By tailoring the output of LLMs according to users' trust attributes, systems can prevent unauthorized access to sensitive information while maintaining usability.

For instance, an LLM featuring ABAC will assess the user attributes and the environmental context prior to generating the response, in a way that sensitive information is only received by the related persons \cite{hu2014abac, zhang2019abacapis}. It will provide better security about compliance with privacy regulations and still maintain the advantages of AI-based interaction.

\subsection{Sensitive Information Management Techniques}
Sensitive information management is crucial for organizations to protect data confidentiality, integrity, and availability. Effective management techniques include data classification frameworks and content filtering with Data Loss Prevention (DLP) methods. These strategies help organizations identify, categorize, and safeguard sensitive data, ensuring compliance with regulations and mitigating the risk of data breaches.

\textbf{Data Classification Frameworks}
Data classification schemes provide a systematic way or methodology of classifying organizational data according to its sensitivity, value, and regulatory requirements \cite{johnson2007embedding}. During the data classification process, an organization can then apply relevant security controls and policies that ensure non-disclosure, unauthorized alteration, unauthorized destruction, and unauthorized disclosure of sensitive information \cite{blakley2001securityriskmanagement}. Data is commonly classified into categories such as public, internal, confidential, and secret, where each classification has unique treatment considerations \cite{nist2008guidemap}.

Data discovery is crucial for organizations to identify and locate sensitive information across various data repositories, ensuring accurate classification \cite{Memon2020}. Major steps in the data classification framework include full inventory and discovery of data to identify and classify all data assets; both structured and unstructured data are described below \cite{symantec2012bestpractices}. In this regard, classification levels are defined based on levels of sensitivity of data and possible business impacts, underpinned by regulatory requirements and organizational policy guidance from \cite{nist2008guidemap, iso2013iso27001}. Lastly, marking of data assets with the correct designations of classification is done using explicit or automated tools for classification \cite{lewis2014dataclassification}.

Policy development establishes the way of handling, storage, transmission, and disposal of information depending on its classification \cite{whitman2011principlesinfosec}. Employee training and sensitization programs will enable the staff to comprehend data classification policies and their role in data protection \cite{albrechtsen2007qualitativeinfosec}. There should be set monitoring and enforcement mechanisms in place while auditing the data handling practices \cite{mcilwraith2016infosecbehaviour}.

Several open-access tools support data classification efforts:
\begin{itemize}
    \item \textbf{Apache Tika:} An open-source toolkit for detecting and extracting metadata and structured text content from various documents \cite{apache2021tika}.
    \item \textbf{ClassifyIT:} An open-source data classification tool that automates the classification and labeling of data based on predefined rules \cite{classifyit2020}.
    \item \textbf{Data Classification Toolkit for Windows Server:} Provided by Microsoft, this toolkit helps organizations identify, classify, and protect data stored on Windows servers \cite{microsoft2012dataclassification}.
\end{itemize}
These tools assist organizations in efficiently implementing data classification frameworks, enhancing their ability to manage sensitive information effectively.

\textbf{Content Filtering and Data Loss Prevention (DLP) Methods}
Content filtering and DLP are two important components in the prevention of sensitive information leakage \cite{hay2008forensics}. While content filtering includes monitoring and controlling the transfer of data in accordance with the set parameters, DLP focuses on the identification, monitoring, and protection of sensitive data at rest, in use, or even in transit through a network to prevent any potential data leak \cite{zuk2009dataloss}.

Content filtering methods are applied to emails, web traffic, instant messaging, and so forth \cite{khan2016cloudsecurity}. Approaches include keyword-matching-based methods, scanning of words or phrases indicating sensitive information \cite{garfinkel2010digitalforensics}, pattern matching using regular expressions to identify data formats such as a credit card number or a social security number \cite{venter2003taxonomyinfosec}, contextual analysis which estimates the context of the data in order to achieve higher detection accuracy \cite{catteddu2009cloudcomputing}, and machine learning algorithms that learn and identify patterns that appear for sensitive data \cite{shabtai2012data}.

DLP solutions furnish overall strategies for the prevention of data leakage. Endpoint DLP works by ensuring that data is well protected at the endpoint, such as not copying to removable media or printing out \cite{tso2015byod}. Network DLP works by monitoring network traffic to prevent unauthorized data transfers via email or web applications, among other protocols \cite{chernyshev2018healthcare}. Cloud DLP extends protection to data stored or processed in cloud environments, addressing challenges such as data proliferation and multi-tenancy \cite{subashini2011cloudissues}.

Several open-source tools are available to implement content filtering and DLP measures:
\begin{itemize}
    \item \textbf{OpenDLP:} An open-source, agent-based DLP tool that scans systems for sensitive data \cite{opendlp2014}.
    \item \textbf{MyDLP:} An open-source DLP solution providing monitoring and protection for data at rest, in motion, and in use \cite{mydlp2021}.
    \item \textbf{ModSecurity:} An open-source web application firewall that can be configured for content filtering to prevent data leakage through web applications \cite{modsecurity2021}.
\end{itemize}

Implementing content filtering and DLP methods presents challenges such as balancing security with usability, managing false positives, and ensuring compliance with privacy regulations \cite{zheleva2012privacysocialnetworks, cavoukian2012privacybydesign}. Effective DLP requires comprehensive policies defining what constitutes sensitive data and acceptable use \cite{whitman2011principlesinfosec, ashford2014dataloss}, regular employee training on data security practices \cite{albrechtsen2007qualitativeinfosec}, continuous monitoring and updates to adapt to evolving threats \cite{kroll2015bigdata}, and integration with other security measures like encryption and access control for a layered defense \cite{whitman2018infosec}.

\begin{table}[h]
\centering
\caption{Open-Source Tools for Data Classification and Data Loss Prevention}
\begin{tabularx}{\textwidth}{|X|X|X|}
\hline
\textbf{Tool} & \textbf{Description} & \textbf{Type} \\ \hline
\textbf{Apache Tika \cite{apache2021tika}} & Toolkit for metadata extraction and content analysis & Data Classification \\ \hline
\textbf{ClassifyIT \cite{classifyit2020}} & Automated data classification and labeling tool & Data Classification \\ \hline
\textbf{Data Classification Toolkit \cite{microsoft2012dataclassification}} & Tool for identifying and protecting data on Windows servers & Data Classification \\ \hline
\textbf{OpenDLP \cite{opendlp2014}} & Agent-based tool for scanning systems for sensitive data & Data Loss Prevention \\ \hline
\textbf{MyDLP \cite{mydlp2021}} & Solution for monitoring and protecting data at rest, in motion, and in use & Data Loss Prevention \\ \hline
\textbf{ModSecurity \cite{modsecurity2021}} & Web application firewall with content filtering capabilities & Content Filtering / DLP \\ \hline
\end{tabularx}
\end{table}

Content filtering and DLP methods are essential for organizations handling large volumes of sensitive data or subject to strict regulatory requirements \cite{hay2008forensics, zuk2009dataloss}. These techniques help prevent accidental or malicious data breaches, protect organizational reputation, and maintain customer trust.

In addition to conventional privacy-preserving techniques like data classification and DLP, Verykios et al. (2004) explored association rule hiding, proposing strategies that ensure sensitive rules are concealed to protect against unauthorized access while maintaining data utility \cite{verykios2004association, verykios2004stateoftheart}. Further, Feretzakis et al. introduced a novel technique using linear Diophantine equations to hide decision tree rules, minimizing data perturbation and preserving decision-making accuracy by adding the fewest possible new instances \cite{feretzakis2018, feretzakis2019}. This method represents a significant improvement over previous approaches.

\subsection{Ethical and Legal Considerations}

The fast development of artificial intelligence, especially large language models, raises apprehension in the area of ethics and law. All these technologies must fit the requirements of privacy regulations and provide ethical norms for respecting the proper dissemination of the technology. Major regulations, like the General Data Protection Regulation (GDPR) and the Health Insurance Portability and Accountability Act (HIPAA), deal with major privacy-related issues, and some of the ethical principles concerning AI are discussed here. Recently, there have been proposals such as the European Union's proposed AI Act.

Privacy laws like GDPR and HIPAA form a cornerstone in the regulation of how AI systems process personal data. The GDPR, issued by the European Union in May 2018 \cite{eu2016gdpr}, is focused on ensuring individuals have control over their personal information while harmonizing the different data protection regulations across EU member states. It enforces requirements that personal data should be processed lawfully, fairly, and transparently. A controller processes personal data only on a legal basis, such as consent, necessity for a contract, or legitimate interest. Individuals have a number of rights in connection with their personal data, including rights of access, correction, erasure—the so-called "right to be forgotten"—restriction of processing, and rights to data portability \cite{voigt2017gdpr}. Furthermore, the principle of "privacy by design" is central to GDPR compliance, ensuring that privacy considerations are embedded in the design of systems and processes from the outset, as articulated by Cavoukian \cite{cavoukian2011privacybydesign}.

In the event of a personal data breach, the GDPR mandates that organizations notify the relevant supervisory authority within 72 hours, as outlined in the guidelines on breach notification provided by the Article 29 Data Protection Working Party \cite{article29dpwp2018breachnotification}. Exceptions to some of these requirements, known as derogations, are outlined in Article 49 of the GDPR, where specific circumstances allow personal data to be transferred outside the EU without additional safeguards, according to the European Data Protection Board \cite{edpb2018derogations}.

Similarly, HIPAA, enacted in 1996 in the United States \cite{hipaa1996}, lays down standards to protect nationally the security of individually identifiable medical records and other personal health information. The act applies to such covered entities as health plans, healthcare clearinghouses, and healthcare providers, including their business associates. The HIPAA Privacy Rule lists the protective conditions for individually identifiable health information, termed protected health information (PHI) \cite{usdhhs2003hipaaprivacyrule}. The Security Rule describes the protective actions that must be in place to safeguard electronic protected health information (e-PHI), including administrative, physical, and technical safeguards \cite{usdhhs2007securitystandards}.

These regulations affect the development and deployment of AI systems, especially large language models dealing with large amounts of personal and sensitive data. Compliance challenges include ensuring data minimization \cite{eu2016gdpr}, obtaining informed consent for data processing \cite{voigt2017gdpr}, and implementing techniques such as anonymization and pseudonymization to protect personal data \cite{dwork2014differentialprivacy, Rieke2020, rocher2019reidentification}. For organizations with an international presence, lawful cross-border data transfer becomes a critical success factor \cite{edpb2018derogations}. Failure to comply can result in hefty fines under GDPR \cite{eu2016gdpr} and civil or criminal penalties under HIPAA \cite{hipaa1996}.

Apart from these legal considerations, ethical issues are central to the responsible development and deployment of artificial intelligence. Ethical frameworks for AI emphasize transparency, fairness, accountability, and respect for human rights. The Asilomar AI Principles \cite{fliprinciples2017asilomar}, established in 2017, identify areas for discussion in research, ethics, and values, as well as long-term considerations. These principles emphasize transparency and the common benefits of AI. The OECD AI Principles \cite{oecd2019aiprinciples} focus on innovative and trustworthy artificial intelligence that respects human rights and democratic principles.

The European Commission's High-Level Expert Group on Artificial Intelligence has issued the Ethics Guidelines for Trustworthy AI \cite{european2019trustworthyai}, which identify seven key areas on which artificial intelligence systems should be built: human agency and oversight, technical robustness and safety, privacy and data governance, transparency, diversity and fairness, societal and environmental well-being, and accountability. Similarly, IBM’s Principles for Trust and Transparency in AI \cite{ibm2018trustprinciples} emphasize purpose, data ownership, transparency, and fairness within AI systems.

Core ethical principles include transparency and explainability, whereby AI systems must explicate the decision-making process to foster trust in these systems \cite{gunning2019xai, miller2019explanation}; fairness and non-discrimination, ensuring that AI does not discriminate and maintains fairness towards all individuals \cite{mehrabi2021biasfairness}; accountability, with mechanisms designed to deal with negative impacts caused by AI \cite{dignum2018ethicsai, raji2019auditing}; privacy and data governance, respecting privacy rights and responsibly handling data \cite{nissenbaum2004contextualintegrity, belanger2011privacy}; and human-centric design, where AI enhances human capacities and preserves human agency \cite{shneiderman2020humancenteredai}.

A notable recent initiative is the proposed AI Act of the European Union, a sweeping regulatory framework for artificial intelligence. The AI Act adopts a risk-based approach, classifying AI systems by the level of risk they pose, ranging from minimal risk to unacceptable risk. Unacceptable uses of AI, such as social scoring by governments and real-time biometric identification in public spaces, are prohibited \cite{ec2021aiactproposal}. High-risk AI systems, including those used in critical infrastructure, education, employment, and law enforcement, are subject to stringent requirements, including conformity assessments, transparency obligations, and human oversight \cite{veale2021euai}.

\begin{table}[H]
\centering
\caption{Key Legal Regulations and Ethical Principles in AI}
\begin{tabularx}{\textwidth}{|X|X|}
\hline
\textbf{Aspect} & \textbf{Description} \\ \hline
\textbf{GDPR \cite{eu2016gdpr, voigt2017gdpr, edpb2018derogations}} & EU regulation focusing on personal data protection, individual rights, data minimization, consent, and data breach notifications. Non-compliance can result in substantial fines. \\ \hline
\textbf{HIPAA \cite{hipaa1996, usdhhs2003hipaaprivacyrule, usdhhs2007securitystandards}} & U.S. law protecting medical records and health information, including privacy and security rules for PHI and e-PHI, with penalties for violations. \\ \hline
\textbf{Ethical AI Frameworks} & Guidelines promoting transparency, fairness, accountability, and human-centric design in AI, including the Asilomar AI Principles \cite{fliprinciples2017asilomar}, OECD AI Principles \cite{oecd2019aiprinciples}, and EU Ethics Guidelines \cite{european2019trustworthyai}. \\ \hline
\textbf{EU AI Act \cite{ec2021aiactproposal, veale2021euai}} & Proposed EU regulation classifying AI systems by risk level, imposing requirements on high-risk AI, prohibiting certain practices, and aiming to ensure trustworthy AI development. \\ \hline
\textbf{Core Ethical Principles} & Transparency and explainability \cite{gunning2019xai, miller2019explanation}, fairness and non-discrimination \cite{mehrabi2021biasfairness}, accountability \cite{dignum2018ethicsai, raji2019auditing}, privacy and data governance \cite{belanger2011privacy, nissenbaum2004contextualintegrity}, human-centric design \cite{shneiderman2020humancenteredai}. \\ \hline
\end{tabularx}
\end{table}

Developers and users of AI, including large language models (LLMs), should observe these ethical principles and legal requirements. Compliance concerns include avoiding bias and ensuring fairness, as LLMs may inherit biases from training data \cite{mehrabi2021biasfairness, bender2021stochasticparrots}; improving explainability to build trust despite the intrinsic complexities of LLMs \cite{gunning2019xai, miller2019explanation}; ensuring privacy by leveraging methods like differential privacy \cite{abadi2016deepprivacy, dwork2014differentialprivacy} and federated learning \cite{yang2019federatedlearning}; setting up accountability mechanisms to address any harm caused by AI systems \cite{raji2019auditing}, and ensuring AI behavior aligns with human values and ethical norms \cite{christiano2017deepreinforcement}.

In conclusion, ethical and legal considerations are paramount in the development and deployment of AI technologies like LLMs. Compliance with regulations such as GDPR and HIPAA ensures the protection of personal data and individual rights. Adhering to ethical AI principles fosters trust, accountability, and societal acceptance of AI systems. As regulatory frameworks like the EU AI Act emerge, organizations must stay informed and proactive in integrating ethical and legal compliance into their AI practices.

\section{Proposed Framework for Embedding Trust Mechanisms in LLMs}

\subsection{Overview of the Framework}

The proposed framework aims to integrate trust mechanisms within the operation of Large Language Models (LLMs) to manage and control sensitive information disclosure dynamically. This approach draws from key concepts in Role-Based Access Control (RBAC), Attribute-Based Access Control (ABAC), and privacy-preserving techniques such as differential privacy. The primary objective is to balance data utility and protection by adjusting the LLM's output based on the trust level assigned to the interacting user or entity. This framework is applicable across various sensitive domains like healthcare, finance, and legal services where privacy and data protection are paramount.

The framework consists of three main components:

\subsubsection{User Trust Profiling}

User Trust Profiling is the foundational component that evaluates and assigns trust levels to users based on a predefined set of attributes. It determines the level of access each user has to sensitive data, enabling dynamic control of the LLM's outputs. Profiling users based on their role, access purpose, and contextual factors ensures that sensitive information is disclosed only to authorized users.

\begin{itemize}
    \item \textbf{User Role:} Based on RBAC principles, user roles (e.g., Administrator, Healthcare Provider, Public User) define access levels \cite{sandhu1996rbacmodels}. For instance, healthcare providers may have higher access to detailed patient information than administrative staff or general users.
    \item \textbf{Purpose of Access:} Drawing from ABAC, access control extends beyond roles to include the specific purpose for which information is requested \cite{hu2014abac}. A user seeking data for public information purposes may receive de-identified summaries, while those requiring data for medical procedures could access full patient records.
    \item \textbf{Contextual Attributes:} Factors such as location, network security, and device type influence the trust level dynamically. Users accessing data from secure networks or trusted devices receive greater access than those on public networks \cite{belanger2011privacy}.
\end{itemize}

The Trust Score adapts in real-time, ensuring the system remains flexible and robust in different contexts.

\subsubsection{Information Sensitivity Detection}

The second component of the framework, Information Sensitivity Detection, involves real-time identification of sensitive information within the model’s outputs. Since LLMs are trained on large datasets, they might unintentionally generate sensitive or confidential data. To mitigate this, multiple techniques are employed to detect sensitive content and prevent unintended disclosure.

\begin{itemize}
    \item \textbf{Named Entity Recognition (NER):} NER systems are deployed to flag personally identifiable information (PII), such as names, addresses, and social security numbers, as well as sensitive data in domains like healthcare \cite{lample2016ner, lehman2021bertclinical}. NER tools detect sensitive entities before they are included in model outputs.
    \item \textbf{Text Classification:} Machine learning models trained on labeled datasets classify text into sensitivity levels (e.g., public, restricted, or confidential) \cite{sun2019bertclassification}. This classification helps prevent sensitive information from being exposed inadvertently.
    \item \textbf{Contextual Analysis:} Beyond individual entities, contextual analysis tools evaluate the broader context of the content to identify potentially sensitive information that may not include explicit PII \cite{catteddu2009cloudcomputing}. For example, a business transaction's details may not contain direct identifiers but still warrant protection due to confidentiality.
\end{itemize}

By combining these techniques, the framework can detect and flag sensitive data, safeguarding outputs from accidental disclosures.

\subsubsection{Adaptive Output Control}

The Adaptive Output Control component dynamically adjusts the LLM's generated outputs according to the user’s trust profile and the sensitivity of the detected information. This approach ensures that the right information is disclosed to the right users without compromising data privacy.

\begin{itemize}
    \item \textbf{Redaction:} For users with lower trust scores, sensitive information identified by the NER or classification systems is redacted from the output \cite{shabtai2012data}. In legal or medical contexts, specific details like names, account numbers, or patient information may be replaced with placeholders.
    \item \textbf{Summarization:} When redaction is not sufficient, the system can summarize sensitive content, providing generalized insights without revealing confidential details \cite{goyal2021flores, shabtai2012data}. For instance, instead of giving precise patient information, a summary of the medical history may be provided.
    \item \textbf{Differential Privacy:} To further protect against data leakage, differential privacy techniques can be employed \cite{abadi2016deepprivacy}. By adding statistical noise to the outputs, differential privacy ensures that even sensitive patterns are obscured while retaining the general utility of the information. This is particularly crucial for safeguarding training data and preventing adversarial attacks \cite{carlini2019secretsharer}.
    \item \textbf{Role-Specific Responses:} Responses are adjusted according to the user’s role. For example, detailed legal or medical information may be provided to professionals (e.g., doctors or lawyers) while generalized data is shared with public users. The role-based approach ensures compliance with regulations like GDPR and HIPAA \cite{eu2016gdpr, hipaa1996}.
\end{itemize}

By dynamically controlling the output, the system ensures that sensitive information is disclosed only to authorized individuals, reducing the risk of data breaches or non-compliance.

\subsection{User Trust Profiling}

The User Trust Profiling component builds on Attribute-Based Access Control (ABAC) principles, defining user access permissions based on a combination of attributes that allow for fine-grained access control. This mechanism ensures that the right users can access appropriate levels of sensitive information depending on their roles, the context of access, and the purpose of interaction. This profiling approach enhances the security of LLM outputs by tailoring responses to users’ trustworthiness.

The key attributes in trust profiling include:

\begin{itemize}
    \item \textbf{User Role:} The role of the user plays a significant part in determining the level of access they should be granted. For instance, a healthcare provider would require detailed patient information, while a general user might only have access to anonymized or summarized data. This role-based approach aligns with the Role-Based Access Control (RBAC) model \cite{sandhu1996rbacmodels, ferraiolo2003rbac}.
    \item \textbf{Purpose of Access:} The purpose of the user's interaction with the system helps define the scope of data disclosure. For example, if the user’s purpose is to provide a medical diagnosis, the system would disclose more detailed data than for a general research query \cite{hu2014abac, yuan2005abac}. The purpose-driven control enhances data protection by aligning the disclosed information with the specific needs of the user.
    \item \textbf{Contextual Factors:} Contextual elements such as location, device security, and network environment further refine the level of access. For instance, users accessing the system from a secure device within a trusted corporate network may receive more detailed responses than those accessing it from a public or untrusted network. \cite{belanger2011privacy, zhang2015privacypreserving}. This dynamic adaptation ensures that sensitive information is only disclosed in secure environments, reducing the likelihood of data leaks.

    \item \textbf{Behavioral Analytics:} The system continuously monitors user interactions, such as access patterns, frequency of requests, and compliance with policies, to refine trust scores over time. The Trust Score is a dynamic, evolving metric that reflects both historical and real-time inputs, allowing for adaptive access control based on changes in user behavior, contextual factors, and compliance with system policies.

\item \textbf{Anomaly Detection:} To enhance security, the framework employs machine learning algorithms to detect unusual activities that might indicate compromised accounts or malicious intent. Algorithms such as:
\begin{itemize}
\item \textbf{Hidden Markov Models (HMM):} HMMs are useful for modeling sequential user behavior, where observed actions depend on hidden states, such as the user’s intent or trustworthiness. Anomalous transitions between states, such as a sudden switch from accessing general information to confidential data, can indicate suspicious behavior \cite{rabiner1989hmm}.
    
\item \textbf{Bayesian Inference:} A probabilistic approach that updates the trust score based on prior knowledge and real-time user actions. Unusual deviations from expected behavior, such as accessing high-sensitivity data with no prior history of doing so, can be flagged as anomalies \cite{gelman2013bayesian}.

\item \textbf{Support Vector Machines (SVM):} SVMs are effective for detecting outliers in user behavior data. By classifying behavior into normal or abnormal categories, SVMs can detect deviations like abnormal access patterns or excessive requests for sensitive information \cite{cortes1995svm}.

\item \textbf{Autoencoders:} For unsupervised anomaly detection, autoencoders reconstruct typical user behaviors. Any significant deviation from expected behavior, resulting in a high reconstruction error, could signal a security threat \cite{hinton2006autoencoder}.

\end{itemize}
\item \textbf{Multi-Factor Authentication (MFA):} Strengthen user identity verification by incorporating MFA methods like biometric verification or one-time passwords.
\item \textbf{Bias Mitigation:} Employ fairness algorithms and conduct regular audits to identify and correct potential biases in trust scoring.

\end{itemize}

These attributes work together within the profiling system to determine a Trust Score for each user. The Trust Score is a dynamic value that evolves based on real-time inputs, ensuring that only users with sufficient authorization receive sensitive information. This process helps maintain a balance between data utility and privacy.

\subsection{Information Sensitivity Detection}

The Information Sensitivity Detection component ensures that sensitive information is not inadvertently disclosed in the LLM's output. Multiple automated techniques are employed to detect and manage sensitive content, providing robust protection in domains such as healthcare, finance, and legal services.

Key techniques and tools for sensitivity detection include:

\begin{itemize}
    \item \textbf{Named Entity Recognition (NER):} NER is a widely used method for detecting sensitive entities such as names, social security numbers, medical identifiers, and other personally identifiable information (PII) in the text \cite{lample2016ner}. Tools like Microsoft Presidio provide an open-source, customizable solution for identifying and anonymizing sensitive information. Microsoft Presidio uses pre-configured recognizers for common entities and allows users to define custom recognizers, making it highly adaptable to specific privacy needs in domains such as healthcare and finance \cite{microsoft2021presidio}. By tagging these entities as sensitive, the system ensures that they are either redacted or appropriately anonymized before being disclosed to users.
    \item \textbf{Contextual Analysis:} In addition to recognizing specific entities, contextual analysis is employed to detect sensitive information embedded within certain contexts. For instance, business transactions or legal discussions may involve sensitive content that is not directly identifiable through NER. Tools such as spaCy and Apache OpenNLP analyze surrounding language structures and patterns to capture implicit sensitive information that may not include direct identifiers but could still pose a confidentiality risk.\cite{spacy2021, apache2021opennlp}. This capability enhances the system’s ability to capture implicit sensitive information that may otherwise go unnoticed.
    \item \textbf{Privacy-Preserving Techniques:} To protect against the risk of sensitive training data being memorized by the model, privacy-preserving techniques like differential privacy are applied. Differential privacy works by adding controlled noise to the data during training, ensuring that no individual data point can be reconstructed from the model’s outputs \cite{abadi2016deepprivacy}. This is crucial for preventing memorization attacks, in which adversaries attempt to extract sensitive information embedded in the model’s outputs \cite{carlini2019secretsharer}. Open-access libraries such as TensorFlow Privacy and PySyft provide implementations of differential privacy, making it easier to integrate these techniques into machine learning models \cite{tensorflow2021privacy}.

    \item \textbf{Domain-Specific Models:} Utilize transformer-based models fine-tuned on datasets relevant to specific domains (e.g., medical records, financial data) to enhance sensitivity detection.
\item \textbf{Continuous Learning and Feedback:} Leverage user feedback and incident reports to continuously retrain models and update detection algorithms, reducing false positives and negatives.
\item \textbf{Regulatory Alignment:} Ensure compliance with regulations like GDPR and HIPAA by integrating continuous sensitivity detection with privacy requirements.

\end{itemize}

These tools and techniques work together to create a comprehensive framework for detecting and safeguarding sensitive information, ensuring compliance with privacy regulations while maintaining the utility of the LLM’s output.

\subsection{Adaptive Output Control}

The Adaptive Output Control component dynamically adjusts the LLM's output based on the user's trust profile and the sensitivity of the information detected by the model. This ensures that only appropriate and necessary information is disclosed, while protecting sensitive data from unauthorized access. Several strategies can be employed to manage the level of information disclosure.

\begin{itemize}
    \item \textbf{Redaction:} For users with lower trust levels, sensitive information can be redacted or anonymized before it is presented in the output. Tools like Microsoft Presidio, mentioned previously for its role in detecting sensitive entities, can also be employed to automatically redact or replace sensitive information with placeholders. Redaction is particularly useful in healthcare and legal domains, where personal data like patient information or legal identifiers must be protected \cite{microsoft2021presidio}. Open-source tools like Apache Tika are also capable of integrating redaction functionalities into the output pipeline \cite{apache2021tika}. 

    \item \textbf{Summarization:} When full disclosure of detailed information is unnecessary or potentially risky, the system can employ summarization techniques to provide high-level overviews while omitting sensitive details. Tools like OpenAI GPT models or BART (Bidirectional and Auto-Regressive Transformers) are commonly used for summarizing text, which helps maintain the utility of the information without revealing sensitive content \cite{lewis2020bart}. Summarization is particularly valuable in cases where users require general insights without access to detailed records, such as in public reports or research abstracts. While redaction ensures sensitive entities are hidden or replaced with placeholders, summarization provides a high-level overview, removing the need for detailed sensitive data.
    \item \textbf{Differential Privacy:} To ensure that sensitive data points are not leaked even in high-trust environments, differential privacy can be applied to the output. By adding controlled noise to the generated responses, differential privacy protects individual data points from being reconstructed from the model’s output \cite{abadi2016deepprivacy}. Libraries like TensorFlow Privacy and PySyft allow developers to implement differential privacy with minimal impact on the model’s overall performance, making them ideal for applications that handle sensitive training data \cite{tensorflow2021privacy}. This technique is critical in preventing adversarial attacks, such as model inversion attacks, where sensitive data can be inferred by analyzing the model’s predictions \cite{carlini2019secretsharer}.

    \item \textbf{Preventive Content Filtering:} Fine-tune the LLM with datasets that exclude sensitive information, reducing the likelihood of generating such content.
\item \textbf{Differential Privacy Implementation:} Apply differential privacy mechanisms during training to prevent the model from memorizing and leaking sensitive data.
\item \textbf{Feedback Loop:} Introduce user feedback mechanisms that allow for reporting inappropriate or sensitive outputs, which can then be used to update filtering rules.

\end{itemize}

These strategies ensure that the framework can adaptively control the amount and type of information disclosed based on the user’s trust profile, effectively balancing data utility and privacy.

\subsection{Framework Overview: Integrating Trust and Privacy Mechanisms in LLMs}
\begin{figure}[H]
    \centering
    \includegraphics[width=\textwidth]{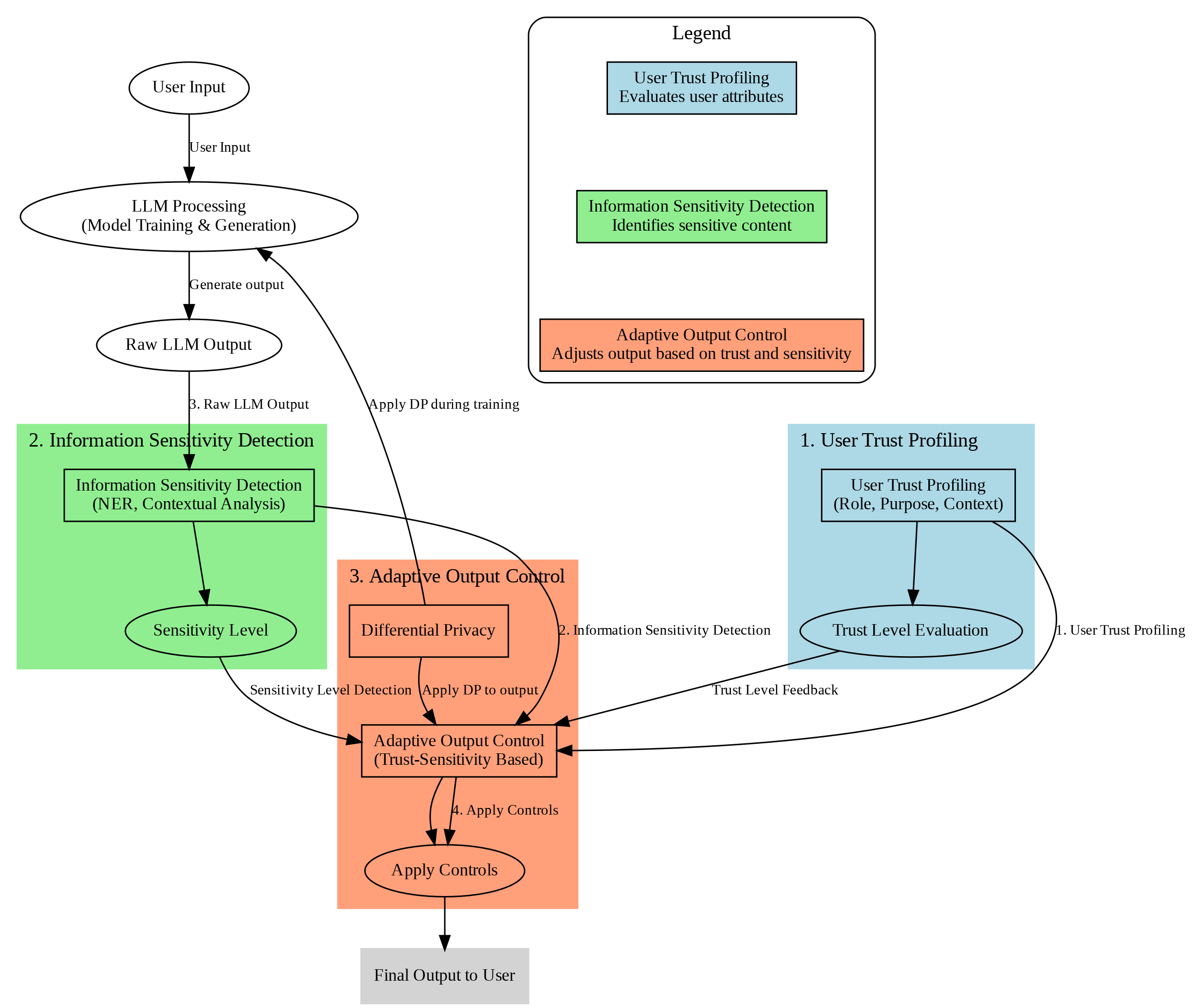}
    \caption{Proposed Framework for Embedding Trust Mechanisms in LLMs}
    \label{fig:flowchart}
\end{figure}

The flowchart provided in Figure \ref{fig:flowchart} outlines a comprehensive framework developed to embed trust mechanisms directly into large language models (LLMs). The aim is to dynamically manage the disclosure of sensitive information while balancing data utility and privacy concerns. This framework is designed specifically for high-stakes environments such as healthcare, finance, and legal sectors where data confidentiality is paramount. By incorporating real-time trust assessments and adaptive output control, the system ensures that the right information is disclosed to the right individuals while minimizing the risk of unauthorized access or data leakage.

The framework revolves around three key components:

\begin{enumerate}
    \item \textbf{User Trust Profiling} (blue section): This module determines the trustworthiness of users by evaluating their role, access purpose, and contextual attributes such as device security and network environment. The system dynamically assigns a trust level score to each user based on these factors, determining the level of access allowed to them. For example, a medical provider accessing sensitive medical data may be assigned a higher trust score compared to a general user accessing public information. This ensures that sensitive information is disclosed only to authorized individuals, while generalized or anonymized outputs are provided to users with lower trust levels.
    
    \item \textbf{Information Sensitivity Detection} (green section): In this module, the system employs techniques such as Named Entity Recognition (NER) and contextual analysis to identify sensitive information within the output generated by the LLM. The sensitivity detected in the process, it informs the type of controls applied in the subsequent stage. Ensuring compliance with the requirements of regulators, including GDPR and HIPAA by flagging personally identifiable information (PII), medical identifiers, and other confidential data for appropriate handling.
    
    \item \textbf{Adaptive Output Control} (orange section): Based on the user’s trust profile and the sensitivity of the detected information, this module applies dynamic controls to the LLM’s output. Such dynamic control may include redaction, summarization, differential privacy, and so on over the output of the LLM. For instance, a low-trust user may receive a summarized or redacted version of the output, while a high-trust user may access more detailed information. This ensures that sensitive data is disclosed appropriately while maintaining the utility of the LLM in diverse scenarios.
\end{enumerate}

The flowchart illustrates how these components interact seamlessly to create a secure and privacy-aware LLM framework. The system dynamically adapts to varying trust levels and information sensitivity, providing a robust solution for deploying LLMs in environments where the protection of sensitive information is critical. Additionally, the framework's flexible design allows for integration with existing access control systems such as Role-Based Access Control (RBAC) and Attribute-Based Access Control (ABAC), further enhancing its applicability in real-world settings.

Figure \ref{fig:flowchart} visually represents the interaction between these components, showcasing the decision-making process that governs how sensitive information is managed. By embedding trust and privacy mechanisms directly into the LLM architecture, the system ensures that information disclosure is both controlled and contextually appropriate, mitigating the risks of data breaches or non-compliance with privacy regulations.

\section{Discussion}

This framework brings together dynamic trust profiling and sensitive information detection with adaptive output control to face the challenges introduced by LLMs in sensitive environments. This section will reflect on how well this framework preserves privacy by using techniques that are aligned with real-world implementation, and what the ethical and technical considerations are to be identified.

\subsection{Dynamic Trust Profiling: Bridging Human and Algorithmic Trust} The most salient feature of the framework is its dynamic trust profiling, which allows flexibility and adaptability in the way it controls access. Drawing inspiration from how human beings assess and adjust trust over time, behavioral analysis, role and purpose evaluation, and contextual influences are infused into a dynamic computation of the user trust by the system. This will be an adaptive approach in high-risk environments such as healthcare, finance, and legal services, where the building of trust will be reassessed continuously based on real-time behaviors and interactions.

While this represents the main features of human trust development, it is still confined to the narrow view brought about by algorithmic trusts. Human trust involves complex emotional and psychological dimensions that cannot be fully captured or quantified by machine learning models.

For example, difficult to include in algorithmic processes are empathy, intuition, and ethical judgment. Models of trust based on only behavioral analytics may lack the nuances of social interaction and even introduce unintended biases into the profiling system.

This necessitates the inclusion of a machine learning-based anomaly detection system, complemented by continuous behavior monitoring, to identify potential threats or malicious activity.

These need to, however, be complemented with fairness algorithms that can reduce the risks of discriminatory practices, since the automated systems usually reproduce the biases encoded in the training data \cite{mehrabi2021biasfairness}. Regular audits and updating of the trust profiling mechanism are indeed critical in ensuring fair and ethical access control in systems that could have implications for individuals' privacy rights or access to basic services.

\subsection{Detection of Information Sensitivity and Privacy Preserving}

This framework incorporates advanced Named Entity Recognition (NER) and contextual analysis techniques to prevent LLMs from disclosing sensitive information. These methods flag personally identifiable information and other sensitive content, but their accuracy is highly dependent on the capabilities of the models to interpret and understand domain-specific data.

As suggested, domain-specific fine-tuning of LLMs offers a fix to this problem by improving the detection of sensitive information in health or finance, for example, where privacy laws like GDPR and HIPAA set high standards on how data should be processed \cite{voigt2017gdpr} \cite{hipaa1996}. Additionally, continuous learning mechanisms and user feedback loops enhance the system's adaptability to evolving data types and privacy requirements. Such features are required to maintain the utility of the model and at the same time ensure compliance with privacy norms.

A key challenge remains to balance privacy preservation with model performance. Techniques like differential privacy protect sensitive data but may reduce model accuracy if not carefully calibrated \cite{abadi2016deepprivacy}. In practice, this still remains one of the open challenges for optimal LLMs with trade-offs between utility of data and privacy, further taking into account that such models in terms of their size and complexity are developing.

\subsection{Adaptive output control: A layered approach to privacy and utility }

The framework includes an adaptive output control mechanism designed to dynamically regulate the disclosure of sensitive information. It integrates redaction, summarization, and differential privacy under the discretion of user profiles so that only the right amount of information is disclosed to a user.

This approach layers real-time trust assessments with sensitivity detection to make granular decisions about information disclosure. Redaction and summarization are crucial when users require insights involving sensitive information but do not need direct access to the details. For example, in healthcare, clinicians will need to see minute details in medical information, while perhaps administrative functionaries only need high-level summaries for performing their duties.

However, this adaptive output control does require the model to be precise in reading user intention and trust levels. This itself is challenging for complex real-world scenarios. The misclassification of the user's trust level or sensitivity of information would lead to either over-restrictive disclosure or overly permissive disclosure. Such scenarios may cause either inefficiency or possible privacy violations \cite{bender2021stochasticparrots}.

Furthermore, differential privacy mechanisms offer strong resistance against data extraction attacks but must be applied carefully to avoid diminishing output utility. In practice, a trade-off can always be expected to be required between protection of privacy and usefulness of shared information, particularly in domains where timely and accurate data is of essence, such as in medical diagnostics or financial decision-making \cite{dwork2014differentialprivacy}.

\subsection{Ethical Considerations and Future Directions}
 As AI systems like LLMs pervade both constructors and users, their creation and deployment have to keep ethical considerations at the forefront of development. This framework addresses potential harms by integrating fairness algorithms, bias mitigation strategies, and regular audits into the trust profiling system. Achieving true fairness in algorithmic trust systems remains challenging due to the inherent biases in the training data and the complexity of human trust. \cite{mehrabi2021biasfairness}.

This is a dynamic and context-aware framework, which innately provides a certain level of transparency into how trust levels are assigned and how information is controlled; however, this is equally as important to make these processes explainable to the users. Further development of the framework could be enhanced by Explainable AI (XAI) techniques, which provide users with insights into how decisions are made, thereby fostering greater trust in the system \cite{gunning2019xai}. Looking ahead, additional empirical testing and refinement of the framework are necessary to surmount limitations for better scalability across domains. Equally, its ability to deal with a decentralized environment will be greatly improved by advanced privacy-preserving techniques such as those proposed in federated learning. This allows for higher privacy and data control with no compromise in the benefits that can be derived from collaborative learning \cite{yang2019federatedlearning}. 

the proposed framework integrates dynamic trust profiling, information sensitivity detection, and adaptive output control to address the challenge of balancing data utility and privacy in LLMs. This is a system that can mimic human-like trust mechanisms by using state-of-the-art privacy-preserving techniques; this represents a new direction for large language model deployments in high-risk environments. Real-world complexity and limitations of algorithmic trust mechanisms call forth continuous research and refinement, nonetheless. Fairness, transparency, and bias mitigation are key ethical considerations that must guide future development. Further, this refinement of mechanisms, with due consideration for user feedback and their own empirical testing, will make this a robust framework with responsible and secure deployment of LLMs in view across sensitive domains.

\section{Conclusion and Future Work}

\subsection{Conclusion}

The present study developed an extended framework, which integrates trust mechanisms into large language models (LLMs) to address one of the most critical challenges in AI—secure and responsible sensitive data processing. While LLMs continue to revolutionize various healthcare, financial, and legal domains by achieving advanced levels of NLP, there has never been a greater need for stringent mechanisms that prevent the exposure of inadvertently revealed private data. The proposed framework provides an innovative solution by integrating User Trust Profiling, Information Sensitivity Detection, and Adaptive Output Control to manage information disclosure securely.

The key contributions of the framework include a User Trust Profiling Module, which profiles users according to their role, intent, and contextual factors, granting access to information. This enables policy-based fine-grained access control of sensitive data through RBAC and ABAC. Unlike standard access control approaches that offer fixed permissions, dynamic profiling ensures that information changes consistently according to the perceived trust levels of the user in real time. This is particularly critical in healthcare, where professionals need immediate access to sensitive information, but privacy concerns must also be addressed.

The Information Sensitivity Detection module enhances the framework’s ability to protect sensitive information. Named Entity Recognition (NER) identifies personally identifiable information (PII), medical identifiers, and other sensitive entities in real time, marking information that requires protection. Contextual analysis goes beyond recognizing named entities, identifying sensitive information embedded in contexts, such as private business contracts or legal documents. These techniques ensure the system remains vigilant in safeguarding information across domains.

The framework also emphasizes privacy-preserving techniques, particularly differential privacy, to mitigate memorization attacks. These attacks exploit the model’s ability to retrieve sensitive information from its training data, constituting a significant privacy risk. Differential privacy introduces noise into the model’s outputs, ensuring that no single data point can be reconstructed, even by highly trusted users. This approach strengthens the system’s defense against adversarial attacks and ensures compliance with privacy regulations like GDPR and HIPAA. The framework effectively balances data utility and privacy while enabling the secure deployment of LLMs in sensitive and regulated environments.

Moreover, the Adaptive Output Control component automatically adjusts the information revealed based on the user’s trust level and the sensitivity of the information. Summarization and differential privacy allow the system to provide tailored responses, ensuring users receive only the necessary information for their tasks without exposing overly sensitive data. For example, a clinician may need detailed patient information, while an administrative staff member may only require a summary. Controlling information granularity based on user roles reduces the risk of breaches and ensures compliance with the principle of least privilege.

The framework leverages open-source tools like Microsoft Presidio, Apache OpenNLP, TensorFlow Privacy, and PySyft to support identification, redaction, and privacy preservation. These tools ensure that the framework is adaptable and scalable across organizations of varying sizes and industries. As these tools continue to evolve, the framework will benefit from their advancements, enhancing its relevance and effectiveness in addressing sensitive information concerns in AI.

In conclusion, the proposed framework presents a robust, adaptable, and scalable solution for embedding trust mechanisms into LLMs. By combining dynamic user profiling, advanced sensitivity detection, and privacy-aware output control, it offers organizations a pathway to harness the power of LLMs while ensuring strict compliance with privacy regulations and the protection of sensitive data. It provides industries with a responsible approach to AI deployment, balancing the trade-off between technological advancement and ethical data handling.

Further refinement and deployment of this framework will enhance its effectiveness, particularly with the integration of new privacy-enhancing technologies and advances in user profiling. As security-critical industries increasingly adopt AI-driven systems, frameworks like the one proposed in this study will play a crucial role in ensuring the safe, ethical, and secure use of these technologies.

\subsection{Future Work}

While the proposed framework provides a solid foundation for integrating trust mechanisms with sensitive information in LLMs, further research is required to assess its scalability and effectiveness across a wide range of real-world contexts. Future work will focus on the actual implementation and testing of the framework in domains such as healthcare, finance, and legal services, where privacy preservation is essential. During this implementation phase, the framework’s ability to meet the complex needs of these high-stakes industries will be better understood.

Much of the future work will involve rigorous empirical testing to evaluate the framework’s security features, scalability, and overall effectiveness. Different domains have unique privacy and data security requirements, which may necessitate adaptations or fine-tuning of the framework. For instance, healthcare settings are bound by HIPAA, while financial sectors adhere to regulations like the Gramm-Leach-Bliley Act. Testing the framework in these environments will provide insights into how it can be tailored for compliance with specific regulatory requirements and highlight any necessary adjustments to enhance security and data protection.

Additionally, future work will focus on balancing data utility and privacy. While the current framework incorporates privacy-preserving strategies like differential privacy and redaction, further research is needed to optimize the trade-off between providing users with useful, actionable information and protecting sensitive data from leakage. This balance is particularly challenging in sensitive environments where even small data leaks can result in significant privacy violations. Advanced testing will help fine-tune the framework to achieve an optimal balance between utility and privacy.

Trust profiling remains an area with room for improvement. While the framework currently combines RBAC and ABAC, future research may explore more adaptive profiling techniques that evaluate user trust more dynamically. Machine learning algorithms could be employed to track user behavior over time, allowing the system to adjust trust levels in real-time based on user activities, such as anomalous behavior indicative of security threats. Multi-factor authentication and persistent monitoring integrated into the trust profile could further enhance security, especially in sensitive environments where unauthorized access must be swiftly detected and mitigated.

Contextual adaptation also requires further research to allow the system to intelligently adjust to a wide range of contextual parameters in real-time, such as device security, network trust, and location. For instance, the level of control for users accessing sensitive data on a protected corporate network may differ from those working remotely on public networks. Developing a dynamic trust level and information disclosure system adaptable to varying contexts without compromising security will be critical to the framework’s success.

Future research will also explore privacy-enhancing techniques like federated learning, which allows LLMs to be trained across decentralized devices without centralizing the data. This approach provides organizations with local control over their data while benefiting from collaborative learning across shared models. However, challenges such as data variability, communication costs, and privacy vulnerabilities in federated learning will need to be addressed. Enhancing federated learning algorithms to handle sensitive data scenarios and incorporating differential privacy will be key areas for future research.

Large-scale deployment will require robust mechanisms for ensuring differential privacy, as the risk of memorization attacks increases with the growing size of LLMs and their training datasets. Future work will focus on developing new algorithms that better balance model performance and privacy. This research will be crucial for ensuring the safe operation of LLMs in sensitive environments without compromising accuracy and utility.

Real-time monitoring and auditing mechanisms will also be necessary to ensure the system operates within set privacy and security standards. Developing automated tools for monitoring and auditing sensitive data interactions will be an important area for innovation, helping organizations stay compliant with evolving privacy regulations such as GDPR, HIPAA, and future AI governance frameworks.

Finally, incorporating explainable AI (XAI) features into the framework will be an important area of future research. As LLMs are increasingly used in sensitive domains, providing users with clear and understandable explanations for system decisions or information disclosures will be critical for maintaining trust. XAI techniques will improve transparency and accountability, helping users and stakeholders better understand the inner workings of the system and ensuring adherence to ethical guidelines.

In summary, while the proposed framework provides a strong foundation for handling sensitive data in LLMs, future work will focus on refining and expanding its capabilities through empirical testing, improved privacy-preserving techniques, enhanced trust profiling, and real-time contextual adaptation. These efforts will ensure that the framework remains relevant and effective in addressing the evolving demands of AI and information privacy.

\end{document}